\documentclass[runningheads]{llncs} 
\usepackage[utf8]{inputenc}
\usepackage{amsmath,amssymb,color,graphicx}
\usepackage[usenames,dvipsnames,svgnames,table]{xcolor}
\usepackage{blkarray}
\usepackage{verbatim}
\usepackage{cite}
\usepackage{adjustbox}

\usepackage{xspace}
\usepackage{caption}
\usepackage{subfig}
\usepackage{hyperref}
\usepackage{wrapfig}
\usepackage{rotating}
\usepackage{todonotes}

\usepackage{amsmath}
\usepackage{algorithm}
\usepackage[noend]{algpseudocode}
\usepackage{mathtools}
\DeclarePairedDelimiter\floor{\lfloor}{\rfloor}

\definecolor{Gray}{gray}{0.85}
\newcolumntype{a}{>{\columncolor{Gray}}l}

\usepackage[misc]{ifsym}  

\usepackage{hyperref}

\usepackage{xcolor}

\title{Symmetry Detection and Classification in Drawings of Graphs}
\author{
Felice De Luca\orcidID{0000-0001-5937-7636}
 \and 
Md Iqbal Hossain(\Letter)\orcidID{0000-0001-6212-7638}
\and 
Stephen Kobourov\orcidID{0000-0002-0477-2724}
}
\institute{
 Department of Computer Science, University of Arizona, USA \email{\{felicedeluca,hossain,kobourov\}@cs.arizona.edu}
 }
  
\begin{document}

\maketitle
\begin{abstract}
Symmetry is a key feature observed in nature (from flowers and leaves, to butterflies and birds) and in human-made objects (from paintings and sculptures, to manufactured objects and architectural design). 
Rotational, translational, and especially reflectional symmetries, are also important in drawings of graphs. Detecting and classifying symmetries can be very useful in algorithms that aim to create symmetric graph drawings and in this paper we present a machine learning approach for these tasks. Specifically, we show that  deep neural networks can be used to detect reflectional symmetries with 92\% accuracy. We also build a multi-class classifier to distinguish between reflectional horizontal, reflectional vertical, rotational, and translational symmetries. Finally, we make available a collection of images of graph drawings with specific symmetric features that can be used in machine learning systems for training, testing and validation purposes. Our datasets, best trained ML models, source code are available online.

\end{abstract}

\section{Introduction}

The surrounding world contains symmetric patterns in objects, animals, plants and celestial bodies. 
A symmetric feature is defined by the repetition of a pattern along one of more axes, called \textit{axes of symmetry}.
Depending on how the repetition occurs the symmetry is classified as \textit{reflection} when the feature is reflected across the reflection axis, and \textit{translation} when the pattern is shifted in the space. 
Special cases of reflection symmetries are horizontal (reflective) symmetry  when the axis of symmetry is horizontal or a vertical (reflective) symmetry when such axis is vertical.
Rotational symmetries occur when the translational axes of symmetry are radial.

Symmetry has been studied in many different fields such as psychology, art, computer vision, and even graph drawing. In psychology, for example, studies on the impact of symmetry on humans show that the vertical symmetry in objects is perceived pre-attentively.
A similar study conducted in the context of graph drawing also shows that the vertical symmetry in drawings of graphs is best perceived among all others~\cite{deluca2018perception}.
In this context, algorithms to measure symmetries in graph drawings have been proposed although it has been shown that these measures do not always agree with what humans perceive as symmetric~\cite{welch2017measuring}.

Convolutional Neural Networks (CNN) have become a standard image classification technique~\cite{krizhevsky2012imagenet}.
CNNs automatically extract features by using information about adjacent pixels to down-sample the image in the first layers, followed by a prediction layer at the end. 

Led by the lack of a reliable way to identify a symmetric layout and eventually classify it by the symmetry it contains, in this paper we consider CNNs for the detection and classification of symmetries in graph drawing. 
Specifically we consider the following two problems: (i) Binary classification of symmetric and non-symmetric layout; and (ii) multi-class classification of symmetric layouts by their type: horizontal, vertical, rotational, translational.
In particular, our contributions are as follows:
\begin{enumerate}
    \item We describe a machine learning model that can be used to determine whether a given drawing of a graph has reflectional symmetry or is not-symmetric (binary classification). This model provides 92\% accuracy on our test dataset.
    \item We describe a multi-class classification model to determine whether a given drawing of a graph has  vertical, horizontal, rotational, or translational symmetry.  This model provides  99\% accuracy on our test dataset. 
    \item We make available training datasets, as well as the algorithms to generate them.
\end{enumerate}
The full version of this paper contains more details, figures and tables~\cite{de2019symmetry}.


\section{Related Work}\label{se:relatedwork}

Symmetry detection has applications in different areas such as computer vision, computer graphics, medical imaging, and robotics. 
Competitions for symmetry detection algorithms have taken place several times; for example, see Liu \textit{et al.}~\cite{liu2013symmetry}. For reflection and translation symmetries the problem can be interpreted as computing one or more axes of symmetry~\cite{kokkinos2006bottomup}.
In the context of graph drawing, symmetry is one of the main aesthetic criteria~\cite{purchase2002metrics}.

\paragraph{Symmetry detection and computer vision: } 
Detection of symmetry is an important subject of study in computer vision~\cite{atallah1985symmetry, liu2010computational, mitra2013symmetry}. The last decades have seen a growing interest in this area although the study of bilateral symmetries in shapes dates back to the 1930s~\cite{birkhoff1932aesthetic}.
The main focus is on the detection of symmetry in real-world 2D or 3D images. 
As Park \emph{et al.}~\cite{park2008performance} point out, although symmetry detection in real-world images has been widely studied it still remains a challenging, unsolved problem in computer vision.
The method proposed by Loy and Eklundh~\cite{loy2006detecting} 
performed best in a competition for symmetry detection~\cite{liu2013symmetry} and is considered a state-of-the-art algorithm for computer vision symmetry detection~\cite{cicconet2016convolutional, park2008performance}. 
Symmetries in 2D points set have also been studied and  Highnam~\cite{highnam1985optimal} proposes an algorithm for discovering mirror symmetries.
More recently, Cicconet \textit{et al.}~\cite{cicconet2016convolutional} proposed a computer vision technique to detect the line of reflection (mirror) symmetry in 2D and the straight segment that divides the symmetric object into its mirror symmetric parts. Their technique outperforms the winner of the 2013 competition~\cite{liu2013symmetry} on single symmetry detection.

\paragraph{Symmetry detection and graphs:}
In graph theory the symmetry of a graphs is known as automorphism~\cite{lubiw1981some} and testing whether a graph has any axial symmetry is an NP-complete problem~\cite{manning1990geometric}.
A mathematical heuristic to detect symmetries in graphs is given in~\cite{fraysseix1999heuristics}. 
Klapaukh~\cite{klapaukh2014empirical,klapaukh2018symmetry} and Purchase~\cite{purchase2002metrics} describe algorithms for measuring the symmetry of a graph drawing. While the first measure analyzes the drawing to find reflection, rotation and translation symmetries, the latter considers only the reflection. 
Welsh and Kobourov~\cite{welch2017measuring} evaluate how well the measures of symmetry agree with human evaluation of symmetry. The results show that in cases where the Klapaukh and Purchase measures strongly disagreed on the scoring of symmetry, human judgment agrees more often with the Purchase metric.

\paragraph{Symmetry detection and machine learning:}
Convolutional neural networks can be a powerful tool for the automatic detection of symmetries.
Vasudevan~\emph{et al.}~\cite{vasudevan2018deep} use this approach for the detection of symmetries in atomically resolved imaging data. The authors train a deep convolutional neural network for symmetry classification using 4000 simulated images, 3 convolutional layers, a fully connected layer, and a final ‘softmax’ output layer on this training dataset. After training over 30 epochs, the authors obtained an accuracy of $85\%$ on the validation set.
Tsogkas and Kokkinos~\cite{tsogkas2012learningbased} propose a learning-based approach to detect symmetry axes in natural images, where the symmetry axes are contours lying in the middle of elongated structures. 
To the best of our knowledge, there are no prior machine learning approaches for detecting or classifying symmetries in graph drawings.


\paragraph{Neural networks for image classification and detection:}
Convolutional Neural Networks (CNNs) are standard in image recognition and classification, object detection, and  video analysis.  The Mark I Perception machine was the first implementation of the perceptron algorithm in 1957 by Rosenblatt~\cite{rosenblatt1958perceptron}.
Widrow and Hoff proposed a mutlilayer perceptron~\cite{widrow199030}. Back-propagation was introduced by Rumelhart \textit{et al.}~\cite{rumelhart1988learning}.  LeNet-5~\cite{lecun1998gradient} was deployed for zip code and digit recognition. 
In 2012, Alex Krizhevsky~\cite{krizhevsky2012imagenet} introduced CNNs with AlexNet. 
Szegedy \textit{et al.}~\cite{ioffe2015batch} introduced GoogLeNet and the Inception module.
Other notable developments include VGGNet~\cite{simonyan2014very} 
and residual networks (ResNet)~\cite{he2016deep}. 

\section{Background and Preliminaries }
\label{se:Background}

In this section we give a brief overview of machine learning in the context of our experiments. We also attempt to clarify some of the terminology we use throughout the paper, focusing in particular on \textit{Deep Neural Networks} and \textit{Convolutional Neural Networks}.


A deep neural network is made of several layers of  neurons. Information flows through a neural network in two ways: via the \textit{feedforward network} and via \textit{backpropagation}.
During the training phase, information is fed into the network via the input units, which trigger the layers of hidden units, and these in turn arrive at the output units. This common design is called a \textit{feedforward network}. Not all units fire all the time. Each unit receives inputs from the units of the previous layer, and the inputs are multiplied by the weights of the connections they travel along. Every unit adds up all the inputs it receives in this way and if the sum exceeds a certain threshold value, the unit fires and triggers the units it is connected to in the next layer.

Importantly, there is a feedback process called \textit{backpropagation} that can be used to improve the weights.
This involves the comparison of the output the network produces with the output it was meant to produce, and using the difference between them to modify the weights of the connections between the units in the network, working from the output units, through the hidden units, and to the input units. 
Over time, backpropagation helps the network to ``learn," reducing the difference between actual and intended outputs.

Convolutional neural network (CNN) are used mainly for image data classification where intermediate layers and computations are a bit different then fully connected  neural networks. Each pixel of input image is mapped with a neuron of the input layer. Output neurons are mapped to target classes. 
Figure~\ref{fig:cnnarch} shows  a simple CNN architecture. Different types of layers in a typical CNN  include:
\begin{figure}
    \centering
    \includegraphics[width=1\textwidth]{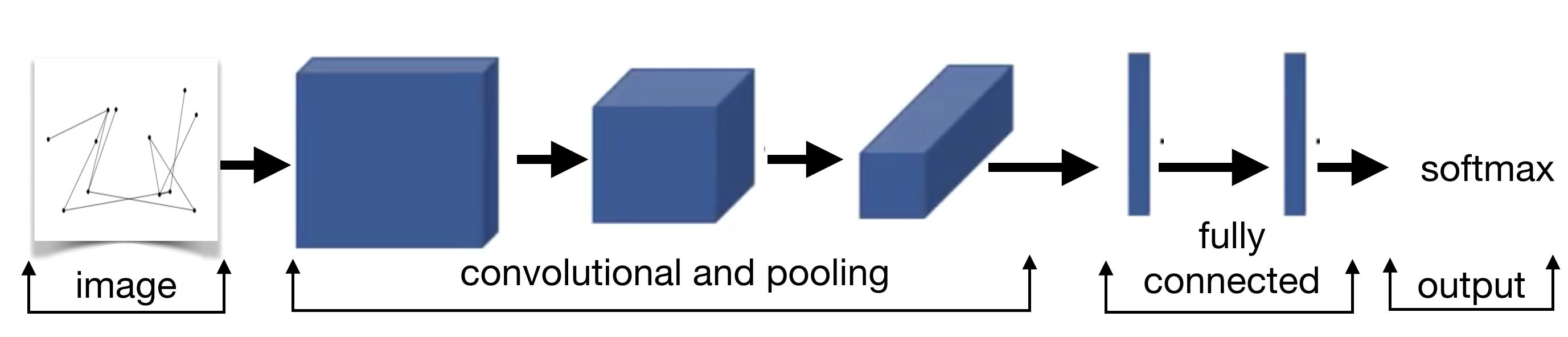}
    \caption{A typical convolutional neural network.}
    \label{fig:cnnarch}
\end{figure}

\begin{itemize}
    \item \textit{convolution layer (convnet)}: in this layer a small filter (usually $3\times 3$) is taken and moved over the image. Applying filters in the layer helps to detect low and high level features in the image so that spatial features are preserved in the layer. The convolutional layer helps to reduce the number of parameters compared to a fully connected layer. Keeping the same set of filters helps to share parameters and sparsity helps to further reduce the parameters. For example, in a $3\times 3$ filter every node in the next layer is only connected to $9$ nodes in the previous layer. This sparse connection helps to avoid over-fitting. 
    \item \textit{activation layer}: this layer applies an activation function from the previous layer. Example functions include ReLU, tanh and sigmoid.
    \item \textit{pooling}: the pooling layer is used to reduce size of the convnet. Filter size $f$, stride $s$, padding $p$ are used as parameters of the pooling layer.  Average pooling or max pooling are the standard options. 
    After applying the pooling to a given image shape ($N_h \times N_w  \times N_c$), it turns into  $\lfloor  \frac{N_h  - f}{s} + 1 \rfloor \times  \lfloor \frac{N_w  - f}{s} + 1 \rfloor \times N_c$. 
 
   
    \item \textit{Fully Connected Layer} (FCL): 
    a fully connected layer creates a complete bipartite graph with the previous layer.  Adding a fully-connected layer is useful when learning  combinations of non-linear features.
    
\end{itemize}
We now review some common machine learning terms. \textit{Training loss} is the error on the training set of data, and \textit{validation loss} is the error after running the validation set of data through the trained network.  Ideally, train loss and validation loss should gradually decrease, and training and validation accuracy should increase over training epochs.
The \textit{training set} is the data used to adjust the weights on the neural network. The \textit{validation set} is used to verify that increase in accuracy over the training data actually yields an increase in accuracy. If the accuracy over the training data set increases, but the accuracy over the validation data decreases, it is a sign of \textit{overfitting}. The \textit{testing set} is used only for testing the final solution in order to confirm the actual predictive power of the network.
A \textit{confusion matrix} is a table summarizing the performance in classification tasks. Each row of the matrix represents the instances in a predicted class while each column represents the instances in an actual class. 
The \textit{precision $p$}  represents how many selected item are relevant and \textit{recall $r$} represents how many relevant items are selected. \textit{$F1$-score} is measured by the formula $2*\frac{r*p}{p+r}$.


\section{Datasets}
\label{se:Dataset}
 
\begin{figure}
    \centering
    \includegraphics[ width=1\textwidth]{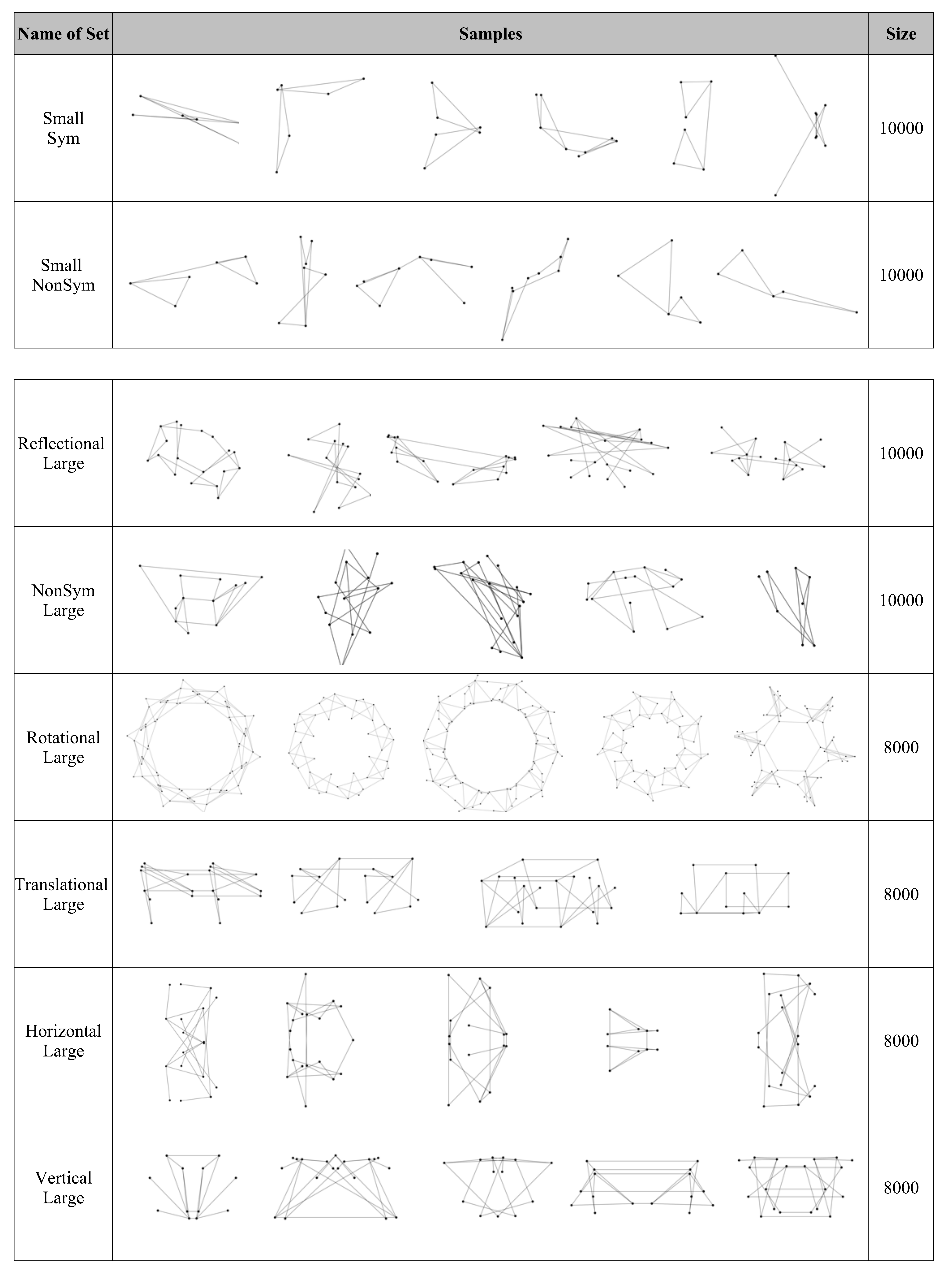}
    \caption{Examples of the different layout instances in our dataset. }
    \label{fig:ourdataset}
\end{figure}

In this section we describe how we generated datasets for our machine learning systems.
To the best of our knowledge, there is no dataset of images suitable for training machine learning systems for symmetry detection in graph drawings. 
Our dataset contains images that feature different types of symmetries, including reflection, translation or rotation symmetries and variants thereof.
An overview all types of layouts is given in Fig.~\ref{fig:ourdataset}. 
%

We started with a dataset of simple symmetric images and inspected the results trying to identify which characteristic of the layout leads to its classification as symmetric or not symmetric. If we observed a characteristic in the symmetric layouts we generated non symmetric layouts that expose it and symmetric layouts without it. Then we 
fed them to the system for the classification. In case of inaccurate results we included the new layouts (that we call \textit{breaking instances} of the dataset) in the training system and repeated the process until we could not identify any other specific feature.

In order to distinguish inputs of different sizes, we refer to layouts in our dataset as \textit{small} or \textit{large} based on the number of vertices, $|V|$.
A small layout has $|V| \in [5, 8]$ while a large layout has $|V| \in [10, 20]$.
The number of edges is a random integer $|E| \in [|V|, \floor*{1.2*|V|}]$. 
The layouts included in the global dataset used for all experiments can be summarized as follows: 
\begin{itemize}
    \item \textit{SmallSym}: small reflective symmetric layout 
    \item \textit{SmallNonSym}: non symmetric generated from SmallSym with random node positions 
    \item \textit{ReflectionalLarge}: large reflective symmetric layouts with random axis of symmetry
    \item \textit{NonSymLarge}:  non symmetric generated from ReflectionalLarge layouts
    \item \textit{HorizontalLarge}: large reflective symmetric layouts with a 0 degree axis of symmetry
    \item \textit{VerticalLarge}: large reflective symmetric layouts with a 90 degree axis of symmetry
    \item \textit{RotationalLarge}: rotational symmetric with random axes between $4$ and $10$ 
    \item \textit{TranslationalLarge}: translational symmetric translated along x-axis 
\end{itemize}

In the remainder of this section we discuss how we generated our layouts and the process that led to them.

\subsection{Reflectional Layout Generation}
A reflectional symmetric layout may expose different characteristics such as ``parallel lines'' orthogonal to the axis of symmetry and edge crossings on the axis of symmetry.

The generation procedure for symmetric graphs and layouts thereof differs slightly depending on the type of symmetry we attempt  to capture. 

We used the procedure for generating a graph and a reflectional symmetric layout with the ``parallel lines'' feature following the algotithm in~\cite{deluca2018perception} as follows. Given a graph with $\frac{n}{2}$ vertices, called a {\em component}, we assign to each vertex of the component positive random coordinates.
Then we copy this component and replace the  $x$-coordinates of each vertex with the negative value of the original. This results in a layout with two disjoint components that are then connected by a random number of edges in $[1, \floor*{|V|/3}]$ selecting random vertices in one component and connecting them to their corresponding vertices in the other component. This results in layouts with vertical axis of symmetry; see Fig.~\ref{fig:layout1}(b). To create layouts with horizontal axis of symmetry we add a 90 degree rotation; see Fig.~\ref{fig:layout1}(a).

The procedure for generating a graph and a reflectional symmetric layout without the ``parallel lines'' feature is described in Algorithm {\em SymGG}.
This algorithm gives an overview on how to create the symmetric versions with the different features.
In the following we explain how we defined {\em SymGG} based on experimental improvements of our dataset.
Given a symmetric graph with $n$ vertices by Algorithm {\em SymGG}, 
we create a non-symmetric layout by assigning to each vertex  of the input graph any random $y$-coordinate and a positive random $x$-coordinate to the vertices with identifier $<\frac{n}{2}$ and a negative random $x$-coordinate, otherwise.

To create reflectional symmetric layouts, instead, if a vertex with identifier $i < \frac{n}{2}$ gets coordinates $(x_r, y_r)$ then the vertex with identifier $i_c = i+\frac{n}{2}$ gets assigned coordinates $(-x_r, y_r)$. 
If the graph has an odd number of vertices then the vertex with identifier $n-1$ gets $x=0$. 
Note that, by construction, the resulting layouts have a vertical axis of symmetry; see Fig.~\ref{fig:layout1}(e).
To create layouts with horizontal axis of symmetry we add a 90 degree rotation; see Fig.~\ref{fig:layout1}(f).

\begin{algorithm}
	\caption*{\textbf{Algorithm SymGG(n,m)}: Symmetric graph generation with $n$ vertices and $m$ edges }\label{alg:symgraphgeneration}
	\begin{algorithmic}[1]
		\State define \textbf{$G = (V, E)$} where $|V|=n$ with id $[0, n-1]$ and $|E|=0$
		\State add $m$ edges to $G$ selecting one or more edge types from [3-6] and continuing with steps [7-12]
		\State for a random edge choose random integers $u, v$ in $[0, n-1]$ such as $(u,v) \notin E$;
		\State for a random edge that does not cross the axis of reflection choose random integers $u, v$ in $[0, \floor*{n/2}-1]$ such as $(u,v) \notin E$;
		\State for parallel edge feature choose random integer $u$ in $[0, \floor*{n/2}-1]$ and $v=u+\floor*{n/2}$ such as $(u,v) \notin E$;
		\State for crossing edge feature choose random integer $u$ in $[0,\floor*{n/2}-1]$ and $v$ in $[n/2, n-1]$ such as $(u,v) \notin E$;
		\State Generate the symmetric edge $(u\_sym, v\_sym)$ of $(u, v)$
		\State $u\_sym = u \mp \floor*{n/2}$ if  $u \gtrless \floor*{n/2}$
		\State $v\_sym = v \mp \floor*{n/2}$ if $v \gtrless \floor*{n/2}$
		\State $u\_sym = u$ if $n$ is odd and $u = n-1$
		\State $v\_sym = v$ if $n$ is odd and $v=n-1$
		\State add $(u, v)$ and $(u\_sym, v\_sym)$ to $E$
	\end{algorithmic}
\end{algorithm}

\subsection{Dataset Definition}
Here we describe the process that led to us to the dataset of reflectional symmetric layouts.

To this aim we generated the SmallSym, SmallNonSym, NonSymLarge and ReflectionalLarge layouts.
\subsubsection{First improvement:}
At first, we trained our system with the reflective symmetric layouts and random layouts generated using the approach in~\cite{deluca2018perception} as described above.

\begin{figure}
\includegraphics[width=1\textwidth]{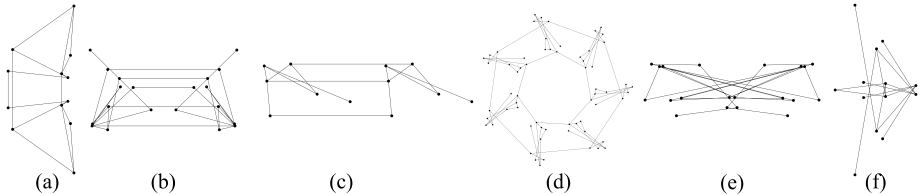}

        \caption{Symmetric layouts in the dataset: (a) Horizontal, (b) Vertical,
        (c) Translational, (d) Rotational,
        (e) Vertical without parallel lines, (f) Horizontal without parallel lines.}\label{fig:layouts}
        \label{fig:layout1}
\end{figure}

\paragraph{Observations: }
Using this simple dataset we observed that the system could always classify the layouts correctly for any of the used layouts.
\paragraph{Layouts characteristic:}
Analyzing the used dataset we observed that the generation algorithm used 
gives symmetric layout for reflective symmetry with a clear symmetric feature that is `parallel lines' orthogonal to the reflection axis. These lines separate two identical but reflected subcomponents, as Fig.~\ref{fig:layout1}(a-b) show. 
\paragraph{Breaking layout:}
After identifying the `parallel lines' feature, we generated non-symmetric layouts with the same feature.
These layouts were created starting from the symmetric layouts and then assigning random positions to the vertices not linked to the parallel edges; an example of random layout with parallel edges is shown in Fig.~\ref{fig:nonsymparallel}. 
Without re-training the system, these layouts are misclassifed as symmetric, breaking the previously built model.

\subsubsection{Second improvement:}

Here we added to our dataset the breaking instances of the previous model and new symmetric layouts that do not show the `parallel lines' feature.
The parallel lines of a symmetric layouts are given by vertices that are connected to their reflected copy (since they share either the $x$ or $y$ coordinate in the space). 
The new layouts we generated have the two subcomponents not only connected by edges between a vertex and his reflected copy 
but also by edges connecting a random vertex of one component to a random vertex of the other (and viceversa to keep the symmetry).
These edges generate crossings on the axis of symmetry of the symmetric layout, instead of the parallel lines.
Pseudocode for the symmetric graph generation Algorithm SymGG (with even number of edges as input) can be found above.

\begin{figure}

\centering
  	\subfloat[~]{\label{fig:nonsymrandom}\includegraphics[width=0.2\textwidth]{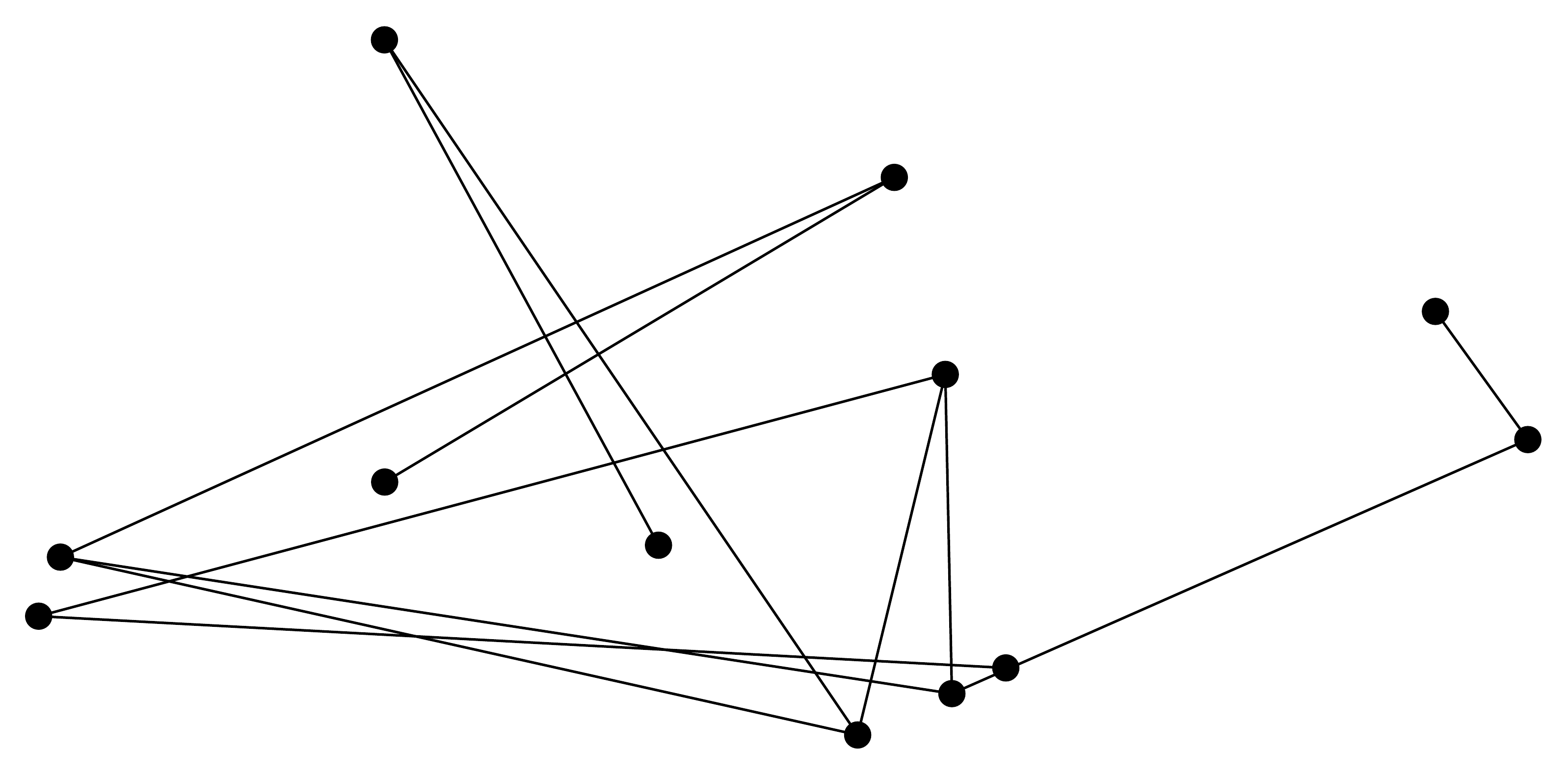}}
  	\subfloat[~]{\label{fig:nonsymparallel}\includegraphics[width=0.2\textwidth]{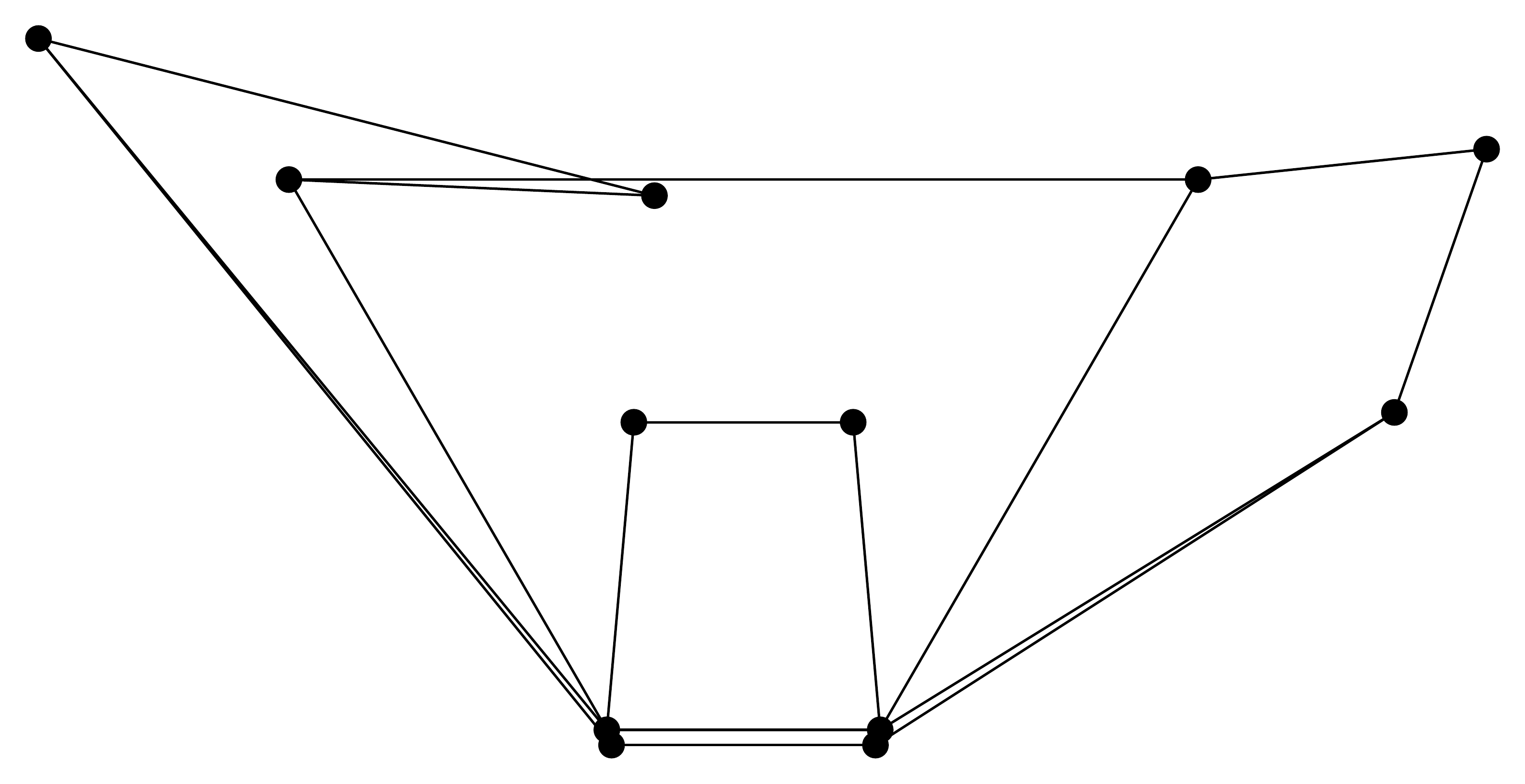}}                
  	  	\subfloat[~]{\label{fig:nonsymcrossings}\includegraphics[width=0.2\textwidth]{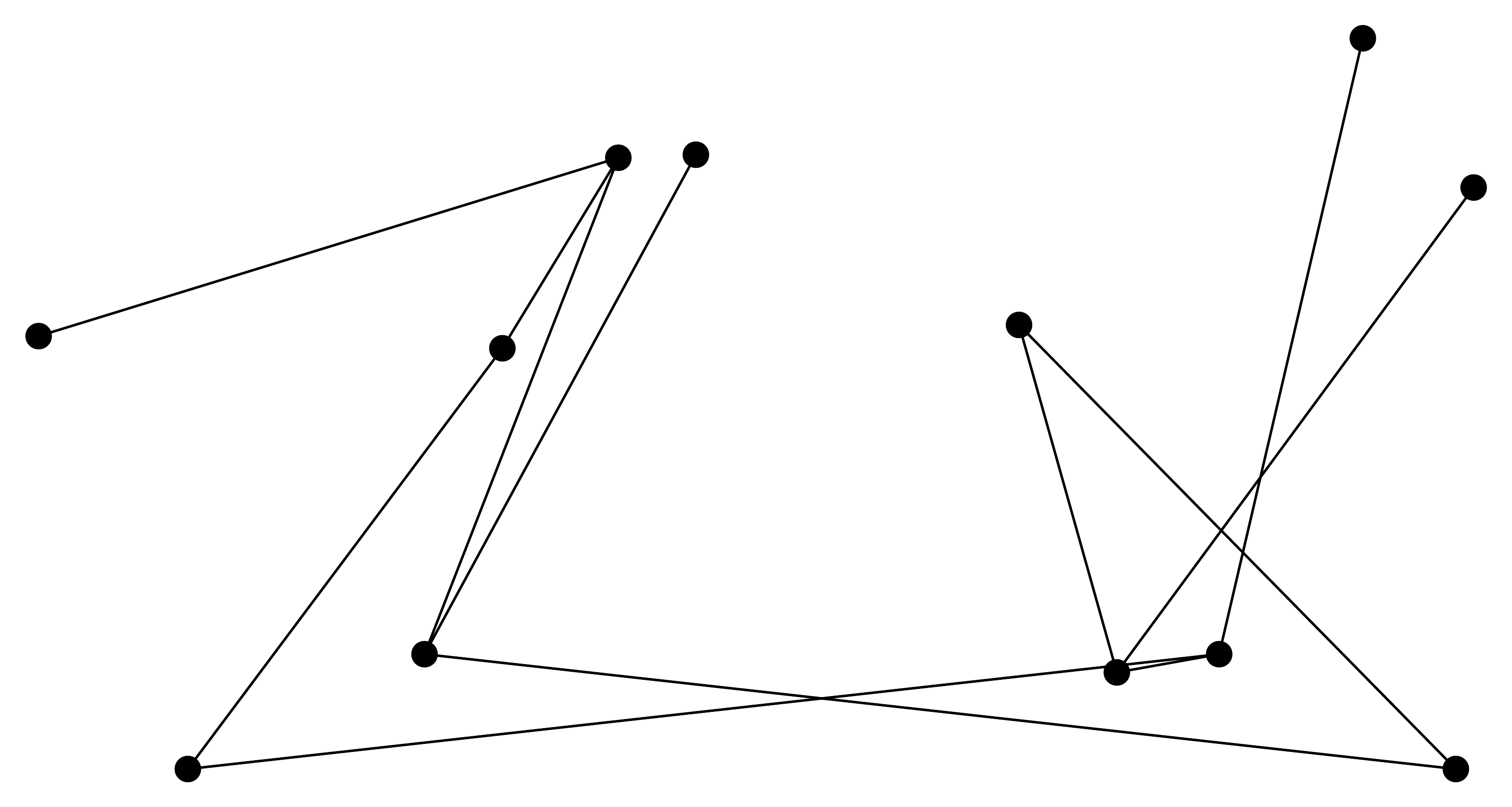}}                
        \caption{Non symmetric layouts in the dataset: (a) Random, (b) with Parallel Lines, (c) with Crossings}\label{fig:nonsymlayouts}
\end{figure}

Analogously we generated some random layouts that show the same feature, starting from a symmetric layout with non-parallel edges and shuffling the position of the vertices not connected to such edges.
Figure~\ref{fig:layout1}(e) illustrates and example of symmetric layout with crossings while Fig.~\ref{fig:nonsymlayouts}(c) depicts a non symmetric layout with crossings.
 
\paragraph{Observations:} 
Training the system with these new layouts we obtained  good results on all layouts, including those misclassified in the previous setup. 

\paragraph{Breaking instance:}
Inspecting the current dataset we identified another characteristic of the current symmetric layouts: an even number of vertices. We then generates symmetric layouts with an odd number of vertices.
The generation algorithm for these layouts is given in Algorithm SymGG. 
Again, without training, the the current system fails on such layouts misclassifying them as a non-symmetric.
Further, we observed that rotating the symmetric layouts also makes our machinery fail. 

\subsubsection{Final improvement:}
Here we added to our dataset instances with odd number of vertices for both symmetric and non symmetric layouts. We also added  instances rotated by a random angle between 0 and 360. Since we could not find further breaking instance for this dataset, we used it for our experiments.

\subsection{Other symmetric layouts}
In addition to the instances above we generated the translational layouts and rotational layouts using the algorithm in~\cite{deluca2018perception}, as follows.

To create translational symmetric layouts we use the same process of generation of reflectional symmetric layout with parallel edges above but instead of taking the negative value of the $x$-coordinate of the copied component we shift each component by a predefined value $\delta$. If a vertex in the given component gets coordinates $(x, y)$ then the vertex in the copied component is assigned coordinates $(x-\delta, y)$; see Fig.~\ref{fig:layout1}(c). 

 The generation process for rotational symmetric layouts  
 is different, since the number of vertices depends on the number of symmetric axes. To generate such layouts we start from a given graph component with $n$ vertices and then 
we select a random number of radial symmetric axes in the range $[4, 10]$. After assigning a random position to the vertices of the component we copy and shift it over the reflection axes. Then we choose two random vertices in the component and use them to connect pairs of rotationally consecutive components; see Fig.~\ref{fig:layout1}(d).

\section{Experimental Setup}\label{sec:experiment} 

Our images are in black and white with a size of $200 \times 200$ pixels.  We use 1 pixel for the edge width and $3 \times 3$ pixels for a vertex. 
  We configured our system with the following settings:  1 grayscale channel, with resealing by 1/255,   batch size 16 and  number of epochs 20.
In all of our experiment we use 80\% of our data as training set, 10\% as validation set, and 10\% as test set. Test sets are never used in during training, those are reserved for computing the final accuracy. During training, in every epoch we check the validation accuracy and save the best trained model as checkpoint. The best trained model is used on the final test set. 

Since images of graph drawings have different features than that of real-world images (e.g., textures and shapes), we tested different popular CNN architectures with same parameter settings. 

         \begin{table}[h!]
         \begin{center}
         \begin{tabular}{ | l  |c |c|c|c| }
         \hline
     
Name & parameters & layers & references & our training time (h) \\ \hline 
ResNet50 & 23.59M & 177  & ~\cite{he2016deep} & 15.25 \\ \hline 
MobileNet & 3.23M & 93 & ~\cite{howard2017mobilenets} & 6.22 \\ \hline 
MobileNetV2 & 2.26M & 157 & ~\cite{sandler2018mobilenetv2} & 8.36 \\ \hline 
NASNetMobile & 4.27M & 771 & ~\cite{zoph2018learning} & 5.79 \\ \hline 
NASNetLarge & 84.93M & 1041 & ~\cite{zoph2018learning} & 10.21 \\ \hline 
VGG16 & 107.01M & 23  & ~\cite{simonyan2014very}& 24.24 \\ \hline 
VGG19 & 112.32M & 26 & ~\cite{chen2017sca} & 25.32\\ \hline 
Xception & 20.87M & 134 &  ~\cite{chollet2017xception} & 19.59\\ \hline 
InceptionResNetV2 & 54.34M & 782 & ~\cite{szegedy2017inception} & 15.18 \\ \hline 
DenseNet121 & 7.04M & 429 & ~\cite{huang2017densely} & 20.11\\ \hline 
DenseNet201 & 18.32M & 709 & ~\cite{huang2017densely} & 28.49 \\ \hline 
 
\end{tabular}
\vspace{1em}
       \caption{Overview of the CNN models used in the experiment.}
      \label{tbl:pic}
      \end{center}
      \end{table}


We use CNN architectures from the Keras implementation;
Keras is a high-level API of Tensorflow that supports training with multiple CPUs\footnote{\url{https://github.com/keras-team/keras/tree/master/keras/applications}}. 
For our experiments, we used the High Performance Computing system at the University of Arizona. Specifically, training was done on 28 CPUs,  each with Intel Xeon 3.2GHz processor and  6GB of memory. 
Training time for the different models ranged from 6 to 29 hours; see Table~\ref{tbl:pic}.

\section{Detecting reflectional symmetry}
\label{se:bc}
\textbf{\textit{Small Binary Classification} (SPBC) Experiment:} 
In this experiment we test how accurately we can distinguish between drawings of graphs with reflectional symmetry and ones without. We use a binary classifier trained on \textit{SmallSym} and \textit{SmallNonSym} instances from our dataset; see Fig.~\ref{fig:ourdataset}. 
We use the \textit{InceptionResNet} CNN model with 12000 images for training, 2000 images for validation, and 2000 image for testing.
The model achieves 92\% accuracy. 
We evaluated several different models before settling on \textit{InceptionResNet}; see 
the full paper for more details~\cite{de2019symmetry}.


We cross-validate our results with two earlier metrics specifically designed to evaluate the symmetry in drawing of graphs, namely the 
Purchase metric~\cite{purchase2002metrics} and the Klapaukh metric~\cite{klapaukh2014empirical}.
These two metrics were not designed for binary classification, but given a graph layout they provide a score in the range $[0, 1]$. We interpret a score of $\ge0.5$ as a vote for ``symmetry" and a score of $<0.5$ as a vote of ``no symmetry."
We can now compare the performance of our CNN model against those of the Purchase metric and the Klapaukh metric on the same set of 2000 test images. We report accuracy, precision, recall and F1-score in Table~\ref{table:HKI}.  
We can see that while the two older metrics perform well, the CNN is better in all aspects (except recall, where the Purchase metric is .01\% better).

\begin{table}[htbp]
\centering
\begin{tabular}{|l|l|l|l|l|}
\hline
Model & Accuracy & Precision & recall & F1-Score \\ \hline
Purchase~\cite{purchase2002metrics} & 82\% &  0.67 & 0.96 & 0.79 \\ \hline
Klapaukh~\cite{klapaukh2014empirical} &  82\% & 0.80 & 0.86& 0.83 \\ \hline
InceptionResNet & 92\% & 0.90  & 0.95 & 0.93  \\ \hline
\end{tabular}
\vspace{1em}
 \caption{Comparison between the CNN model and existing symmetry metrics.}
\label{table:HKI}
\end{table}

Training loss, validation loss, training accuracy and validation accuracy for our {\textit Experiment SPBC} are shown 
in the full version of the paper~\cite{de2019symmetry}.

\section{Detecting different types of symmetries}\label{se:mc}
\textbf{\textit{Multi-class symmetric layouts classification (LHVRT) Experiement:}}
In this experiment we test how accurately we can distinguish between drawings of graphs with different types of symmetries. We use a multi-class classifier trained on several types of symmetries: Horizontal, Vertical, Rotational and Translational.
Recall that Horizontal and Vertical are special cases of reflection symmetry, where the axis of reflection is horizontal or vertical, respectively.

We train the CNN with  \textit{HorizontalLarge}, \textit{VerticalLarge}, \textit{RotationalLarge}, and \textit{TranslationalLarge} instances from our dataset; see Fig.~\ref{fig:ourdataset}.

We use the \textit{ResNet50} CNN model with 16000 images for training, 2000 images for validation, and 4280 image for testing.
The model achieves 99\% accuracy. Table~\ref{table:CM} shows the corresponding confusion matrix. 
We evaluated several different models before settling on \textit{ResNet50}.
Training loss, validation loss, training accuracy and validation accuracy for our {\textit Experiment LHVRT} are shown in the full version of the paper, where more results and  discussion thereof can also be found~\cite{de2019symmetry}.

\begin{table}[]
\centering
\begin{tabular}{|l|l|l|l|l|}
\hline 
 & HorizontalLarge &  RotationalLarge &  TranslationalLarge &  VerticalLarge\\
\hline 

 HorizontalLarge & 1280&    0&    0&    0\\ \hline
 RotationalLarge &          0& 800&    0 &    0\\ \hline
  TranslationalLarge &        0&    0&  798&    2\\ \hline
   VerticalLarge &       0&    0&    1& 1599\\ \hline
\end{tabular}
\vspace{1em}
\caption{Confusion matrix from \textit{ResNet50}. Each row of the matrix represents the instances in a predicted class while each column represents the instances in an actual class.}
\label{table:CM}
\end{table}


\section{Conclusions}

In the experiments above we achieved high accuracy for both detection and classification. 
Compared to existing evaluation metrics for symmetric layout we observed that our machinery outperforms the mathematical formulae proposed when used as classifiers.

Note, however, that there are many limitations to consider. First of all, we generated all the datasets and have not tested the models on layouts obtained from other layout algorithms. Further, the graphs we used are small and we have not confirmed how well humans agree with the decisions of the machine learning system. Finally, the two tasks we performed are limited in power, and we do not yet have a model that can accurately predict whether a graph drawing is symmetric or not, or which of two drawings of the same graph is more symmetric. 

Nevertheless, we believe our dataset can be useful for future experiments and our initial results on limited tasks indicate that a machine learning framework can be useful for symmetry detection and classification. Our dataset, models, results details can be found in \url{https://github.com/enggiqbal/mlsymmetric}

\medskip
\noindent {\bf Acknowledgement}\\
This work is supported in part by NSF grants CCF-1740858, CCF-1712119, DMS-1839274, and  DMS-1839307. This experiment uses High Performance Computing resources supported by the University of Arizona TRIF, UITS, and RDI and maintained by the University of Arizona Research Technologies department.



\bibliography{symbib}
\bibliographystyle{splncs04}


\newpage
\section*{Appendix}

\section{Discussion}
We performed further experiments to test the behavior of the machine learning system on slightly larger graphs and with slightly more difficult tasks. 
We report on two such  experiments, LNBC and LRefRotTra, below.
 
\subsection{Experiment LNBC (\textit{ReflectionalLarge}, \textit{NonSymLarge})}
In this experiment we use more complex input instances when detecting symmetries. The dataset  \textit{ReflectionalLarge} and \textit{NonSymLarge} are used as symmetric and non-symmetric in this experiment with 10000 samples; see Fig.~\ref{fig:ourdataset}. We use 80\% of our data as training set, 10\% as validation set, and 10\% as test set.  Note that  \textit{ReflectionalLarge} dataset combines all types of reflection symmetries, including horizontal reflection, vertical reflection and arbitrary axis reflection. This makes the set of symmetric instances more varied then when considering only one type of symmetry. Further, \textit{NonSymLarge} contains more complex  non-symmetric instances, where starting from a symmetric layout a few vertices are slightly perturbed in order to break the symmetry. This task is clearly harder and  accuracy decreases to 78\%; see Table~\ref{tbl:totalresults}. 

The experiment above motivates three more focused experiments that we use to identify the nature of the difficulty of the LNBC task . 
\begin{itemize}
    \item {\bf Experiment LHnonSym}: tests whether the model can distinguish between complex non-symmetric (dataset \textit{NonSymLarge}) and only  horizontal symmetric samples (dataset \textit{HorizontalLarge}).
    \item {\bf Experiment LVnonSym}: tests whether the model can distinguish between complex non-symmetric (dataset \textit{NonSymLarge} )  and only  vertical symmetric samples (dataset \textit{VerticalLarge}).  
    \item {\bf Experiment LHVSym}: tests  whether the model can distinguish between horizontal symmetric (dataset \textit{HorizontalLarge}) and only  vertical symmetric samples (dataset \textit{VerticalLarge}).
\end{itemize}
Table~\ref{tbl:totalresults} shows that several of models achieve 100\% accuracy for all 3 of these experiments (LHnonSym, LVnonSym and LHVSym). This provides a possible  explanation for the low accuracy of the LNBC experiment: the machine learning algorithms are struggling to distinguish symmetric from non-symmetric layouts when both the symmetric instances are more complex (different types of symmetries) and when the non-symmetric instances are also more complex (different types of non-symmetric layouts).

\subsection{Experiment LRefRotTra (\textit{ReflectionalLarge}, \textit{RotationalLarge},  \textit{TranslationalLarge})}
We next conducted an experiment to detect the type of symmetry in a given layout, from the possible options: reflectional, rotational, and translational. 
Note that in this experiment we do not distinguish among the various types of reflectional symmetry (horizontal, vertical, arbitrary axis). That is, the reflectional layouts include vertical, horizontal, and reflectional with random angle of rotation samples.  From the total of 24720 instances in these three datasets, we choose 80\%  for training, 10\% for  validation, and 10\% for testing; see the \textit{ReflectionalLarge}, \textit{RotationalLarge}  and \textit{TranslationalLarge} rows in Fig.~\ref{fig:ourdataset}. 


The best performing models achieve 69\% accuracy, which indicates difficulty in distinguishing the different types of symmetries. 
In particular, the confusion matrix in Table~\ref{table:CM2} shows that the translational symmetric instances are incorrectly detected as reflectional symmetric instances.

\begin{table}[]
\centering
\begin{tabular}{|l|l|l|l|}
\hline 
 & ReflectionalLarge &  RotationalLarge &  TranslationalLarge \\
\hline 

 ReflectionalLarge & 872&    0&    0\\ \hline
 RotationalLarge &          0& 800&     0\\ \hline
  TranslationalLarge &        800&    0&  0\\ \hline

\end{tabular}
\vspace{1em}
\caption{Confusion matrix of {\it Experiment LRefRotTra}. Each row of the matrix represents the instances in a predicted class while each column represents the instances in an actual class. Note that the translational symmetric instances are incorrectly detected as reflectional symmetric instances.}
\label{table:CM2}
\end{table}

\section{Experimental statistics }

\begin{figure}
    \centering
     \includegraphics[width=1\textwidth]{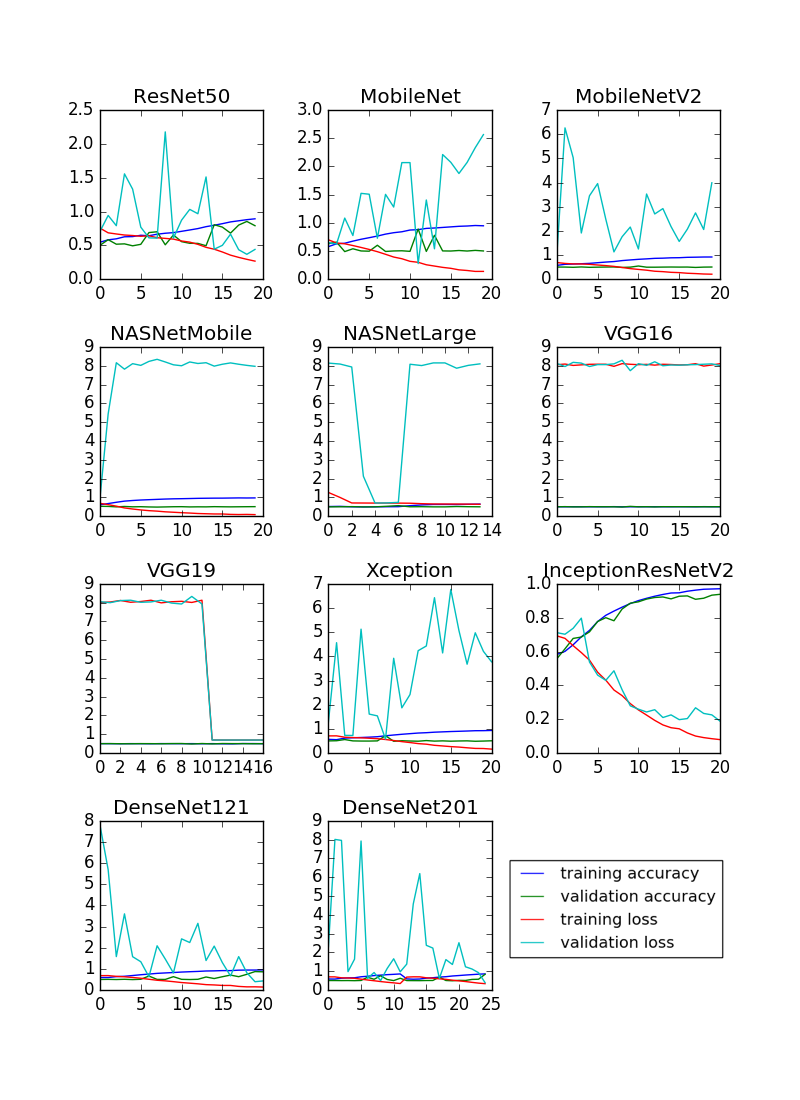}
    \caption{Training loss, validation loss, training accuracy and validation accuracy of different CCN architectures for binary classification. The $x$-axis represents epochs and the $y$-axis represents loss or accuracy.  InceptionResNetV2 shows the correct behavior with decreasing losses increasing accuracy.}
    \label{fig:binary_total}
\end{figure}

\begin{figure}
    \centering
    \includegraphics[width=1\textwidth] {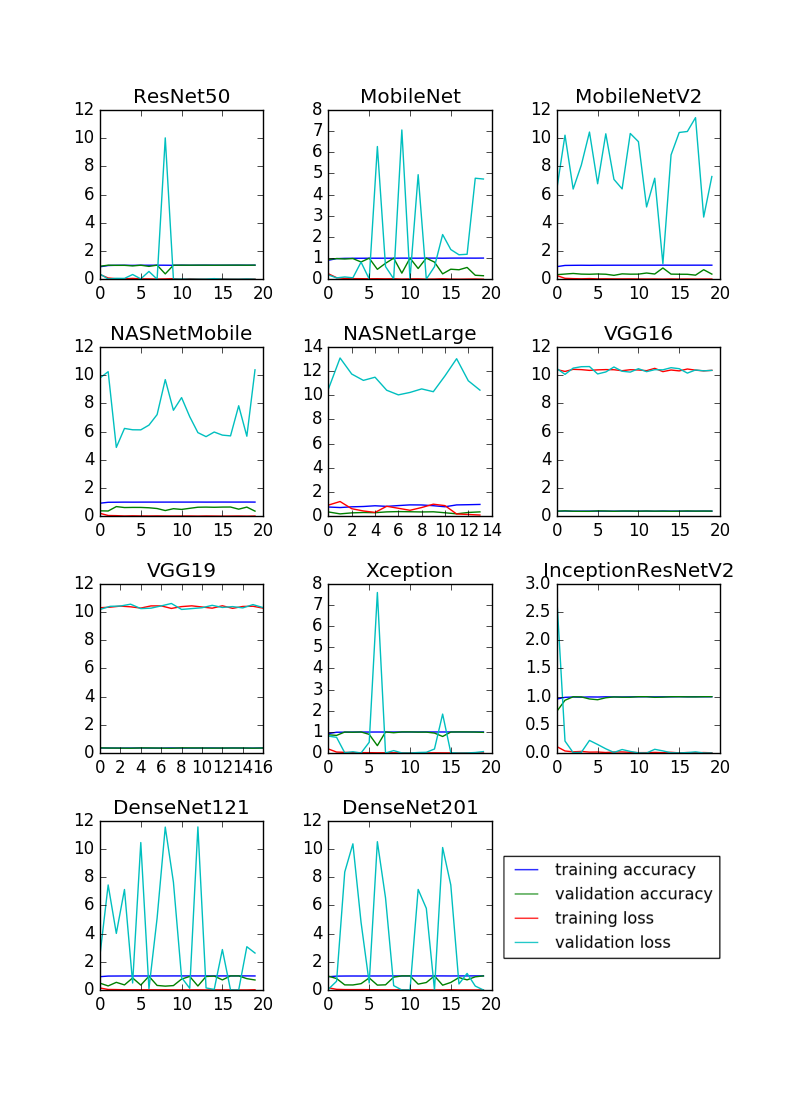}
    \caption{Training loss, validation loss, training accuracy and validation accuracy of different CCN architectures for multi-class classification.}
    \label{fig:multi}
\end{figure}

In this section we present some statistics of training progress of different models. Figure~\ref{fig:binary_total} show training loss, validation loss, training accuracy and validation accuracy of different CCN architectures for \textit{Experiment SPBC}.
For each graphic, the $x$-axis represents the epochs and the $y$-axis represents the value of loss or accuracy, depending on the color of the line (see legend).
Overall, we ovserve that InceptionResNetV2 has the best behavior showing decreasing loss and increasing accuracy. Figure~\ref{fig:multi} shows similar statistics for \textit{Experiment LHVRT} where only two models converge fast, namely 
ResNet50, InceptionResNetV2.

We summarize results of all experiments in the Table~\ref{tbl:totalresults} where training and validation accuracy are reported.

    \begin{table}[htbp]
    \tiny
         \begin{center}
         \begin{tabular}{ | l  |c |a|  c  |a |c| a  |c |a|c|a|c|a|c|a|}
         \hline
model & 
 \multicolumn{2}{|c|}{SPBC} &  \multicolumn{2}{|c|}{LNBC} &  \multicolumn{2}{|c|}{LHnonSym} &  \multicolumn{2}{|c|}{LHVSym} &  \multicolumn{2}{|c|}{LVnonSym} &  \multicolumn{2}{|c|}{LHVRT} &  \multicolumn{2}{|c|}{LRefRotTra} \\
         \hline
     & tracc & vacc &  tracc & vacc &tracc & vacc &tracc & vacc &tracc & vacc &tracc & vacc &tracc & vacc \\ \hline

  ResNet50 & 0.89 & 0.85 & 0.89 & 0.76 & 1.0 & 1.0 & 1.0 & 1.0 & 1.0 & 1.0 & 1.0 & 1.0 & 1.0 & 0.68 \\ \hline
MobileNet & 0.95 & 0.89 & 0.96 & 0.72 & 1.0 & 1.0 & 1.0 & 1.0 & 1.0 & 1.0 & 1.0 & 1.0 & 1.0 & 0.69\\ \hline
MobileNetV2 & 0.92 & 0.55 & 0.93 & 0.42 & 1.0 & 1.0 & 1.0 & 1.0 & 1.0 & 0.89 & 1.0 & 0.8 & 1.0 & 0.68\\ \hline
NASNetMobile & 0.97 & 0.52 & 0.98 & 0.63 & 1.0 & 0.59 & 1.0 & 1.0 & 1.0 & 0.53 & 1.0 & 0.68 & 1.0 & 0.66\\ \hline
NASNetLarge & 0.64 & 0.56 & 0.87 & 0.67 & 0.99 & 0.68 & 1.0 & 0.92 & 1.0 & 0.54 & 0.97 & 0.38 & 1.0 & 0.67\\ \hline
VGG16 & 0.51 & 0.52 & 0.47 & 0.41 & 0.56 & 0.58 & 0.45 & 0.47 & 0.51 & 0.53 & 0.36 & 0.38 & 0.51 & 0.33\\ \hline
VGG19 & 0.51 & 0.51 & 0.46 & 0.42 & 0.56 & 0.58 & 0.45 & 0.46 & 0.51 & 0.53 & 0.36 & 0.37 & 0.51 & 0.34\\ \hline
Xception & 0.94 & 0.72 & 0.97 & 0.71 & 1.0 & 1.0 & 1.0 & 1.0 & 1.0 & 1.0 & 1.0 & 1.0 & 1.0 & 0.68\\ \hline
InceptionResNetV2 & 0.97 & 0.94 & 0.99 & 0.78 & 1.0 & 1.0 & 1.0 & 1.0 & 1.0 & 1.0 & 1.0 & 1.0 & 1.0 & 0.69\\ \hline
DenseNet121 & 0.95 & 0.87 & 0.96 & 0.62 & 1.0 & 1.0 & 1.0 & 1.0 & 1.0 & 1.0 & 1.0 & 1.0 & 1.0 & 0.69\\ \hline
DenseNet201 & 0.86 & 0.83 & 0.89 & 0.69 & 1.0 & 0.99 & 1.0 & 1.0 & 1.0 & 1.0 & 1.0 & 1.0 & 1.0 & 0.68\\ \hline

\end{tabular}
\vspace{1em}
      \caption{Training accuracy (tracc) and validation accuracy (vacc) achieved by different models for the different tasks, at a glance.}
      \label{tbl:totalresults}
      \end{center}
      \end{table}

\end{document}